# On Fusing ChatGPT and Ensemble Learning in Discontinuous Named Entity Recognition in Health Corpora[1]

**Tzu-Chieh Chen [1] and Wen-Yang Lin [2,*]**

[1] Dept. of Computer Science and Information Engineering, National University of Kaohsiung, Taiwan; nick82067@gmail.com
[2] Dept. of Computer Science and Information Engineering, National University of Kaohsiung, Taiwan; wylin@nuk.edu.tw
* Correspondence: wylin@nuk.edu.tw; Tel.: +886-7-5919517

**Abstract:** Named Entity Recognition (NER) has traditionally been a key task in natural language processing (NLP), aiming to identify and extract important terms from unstructured text data. However, a notable challenge for contemporary deep-learning NER models has been identifying discontinuous entities, which are often fragmented within the text. To date, methods to address Discontinuous Named Entity Recognition (DNER) have not been explored using ensemble learning to the best of our knowledge. Furthermore, the rise of large language models (LLMs, such as ChatGPT) in recent years has shown significant effectiveness across many NLP tasks. Most existing approaches, however, have primarily utilized ChatGPT as a problem-solving tool rather than exploring its potential as an integrative element within ensemble learning algorithms. In this study, we investigated the integration of ChatGPT as an arbitrator within an ensemble method, aiming to enhance performance on DNER tasks. Our method combines five state-of-the-art (SOTA) NER models with ChatGPT using custom prompt engineering to assess the robustness and generalization capabilities of the ensemble algorithm. We conducted experiments on three benchmark medical datasets, comparing our method against the five SOTA models, individual applications of GPT-3.5 and GPT-4, and a voting ensemble method. The results indicate that our proposed fusion of ChatGPT with the ensemble learning algorithm outperforms the SOTA results in the CADEC, ShARe13, and ShARe14 datasets, achieving improvements in F1-score of approximately 1.13%, 0.54%, and 0.67%, respectively. Compared to the voting ensemble method, our approach achieved improvements of about 0.63%, 0.32%, and 0.09%. Furthermore, compared to GPT-3.5 and GPT-4, our average results were approximately 7.42%, 0.89%, and 0.54% higher. The results demonstrate the effectiveness of our proposed fusion method of ChatGPT and ensemble algorithms, showcasing its potential to enhance NLP applications in the healthcare domain.

## 1. Introduction

NER tasks have long been fundamental tasks in natural language processing, involving the identification and extraction of important terms from unstructured textual data. Early NER methods often relied on predefined syntactic or semantic rules and relied on annotated corpora, which required significant human resources and time costs for manual analysis [1]. Over the years, approaches that utilize machine learning and deep learning technologies have gained widespread popularity [2]. Most traditional NER models [3]-[6] transform the NER problem into a sequence labeling task [7],[8], where each token is labeled as "B", "I", or "O". Here, "B" indicates the start of an entity, "I" signifies that a token is inside an entity, and "O" marks a token as outside of any entity. As shown in Figure 1, the word "aching" is the beginning of the ADE entity "aching in legs", thus it is labeled as "B", and the words "in" and "legs" are in the inside of the entity, hence labeled as "I". However, this approach of assigning only one label per token can only handle continuous NER tasks and cannot address issues involving irregularly overlapping or discontinuous entities.

[1] A short version of this paper entitled "Beyond GPT-NER: ChatGPT as Ensemble Arbitrator for Discontinuous Named Entity Recognition in Health Corpora" has been accepted for presentation at MedInfo2025.

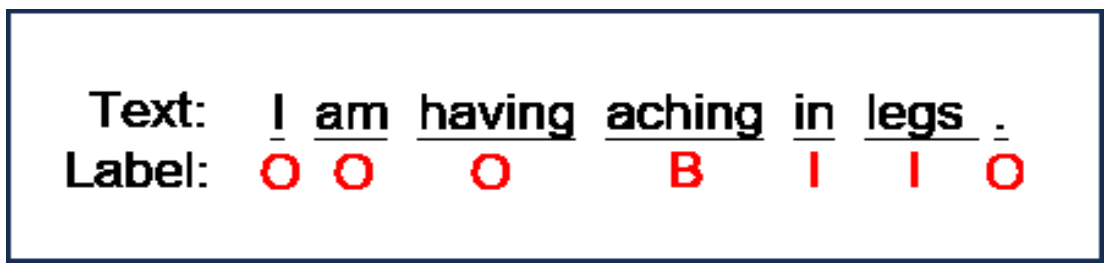

**Figure 1.** Sequence labeling representation method of traditional NER model.

These irregular entities frequently appear in many practical scenarios, particularly in healthcare [9]. In clinical settings, electronic health records typically include valuable information such as medications, symptoms, and ADEs [10], thereby supporting various downstream NLP applications (e.g., question answering [11] and relation extraction [12]). In Figure 2, we illustrate examples of continuous entities and irregular entities. Example 1 shows a sentence with a continuous entity, while Example 2 demonstrates a sentence with two discontinuous entities, including one overlapping entity. Here, "aching in legs" (E1) constitutes a continuous entity, whereas "muscle fatigue" (E2) and "muscle pain" (E3) are discontinuous entities with an overlapping entity, "muscle." These irregular entities formed by non-contiguous spans pose unique complexities in extraction due to their dispersed occurrences in textual data, presenting more significant challenges than traditional NER tasks.

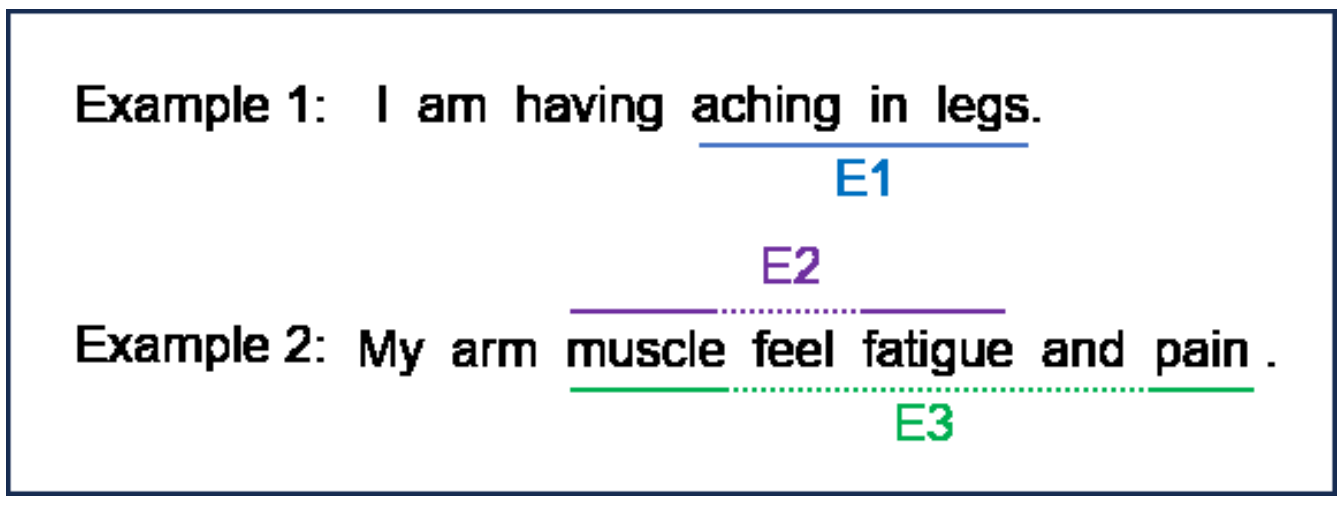

**Figure 2.** Examples of continuous entities and irregular entities.

In recent years, several methods have been proposed [13]-[16] to address the challenges of sequence labeling tasks in DNER. While some research has attempted to solve NER problems through ensemble learning [17], to the best of our knowledge, no one has proposed using ensemble learning to tackle DNER issues. In November 2022, OpenAI developed a large language model, GPT-3.5 [18], capable of understanding human natural language and responding meaningfully. ChatGPT is trained on extensive data and deep learning algorithms, enabling it to perform tasks such as conversation, question answering, and translation [19]. Numerous studies have demonstrated ChatGPT's significant effectiveness across various NLP tasks [20]. In March 2023, OpenAI released GPT-4, which boasts even more training data and enhanced capabilities compared to GPT-3.5 [21]. However, we find that most people primarily use ChatGPT as a tool for problem-solving [22], without exploring its potential as an arbiter in integrating ensemble learning algorithms to address NLP challenges.

In this context, our research explores a novel integrated learning framework for DNER, leveraging ChatGPT as an arbiter and integrating multiple deep learning models for DNER processing through custom prompt engineering. We aim to enhance the model's ability to effectively recognize discontinuous entities. We believe that ensemble learning techniques provide robustness and generalization capabilities, enabling our approach to surpass traditional single-model and standalone ChatGPT approaches. To evaluate the effectiveness of our proposed method, we conducted comprehensive experiments on CADEC [23], ShARe13 [24], and ShARe14 [25] datasets. Our evaluation not only compared the performance of our method with five baseline methods but also included individual evaluations with GPT-3.5, GPT-4, and hard voting ensemble methods. Experimental results demonstrate that our method significantly outperforms the state-of-the-art (SOTA) results by approximately 1.13%, 0.54%, and 0.67% in terms of F1-score on the CADEC, ShARe13, and ShARe14 benchmark datasets, respectively, and by approximately 0.63%, 0.32%, and 0.09% compared to the results of the hard voting ensemble method. Furthermore, compared to GPT-3.5 and GPT-4, our average results were approximately 7.42%, 0.89%, and 0.54% higher. The results demonstrate the effectiveness of our proposed fusion method of ChatGPT and ensemble algorithms, showcasing its potential to enhance NLP applications in the healthcare domain.

## 2. Related Work

### *2.1. Prompt Engineering*

Prompt engineering refers to the deliberate design and adjustment of input prompts in generative AI systems, such as ChatGPT, to guide the model towards generating desired outputs, thereby enhancing performance on specific tasks [26]. In recent years, several studies have indicated that prompt engineering has a crucial impact on the performance of generative AI [27],[28]. Moreover, in the medical domain, researchers have explored prompt engineering for tasks like medical question answering [29] and medical NER [30] using ChatGPT. Effective prompt engineering maximizes the quality of model outputs, ensuring relevance and accuracy in generated content while minimizing irrelevant or biased results.

### *2.2. Baseline Deep Learning DNER Models*

In our research approach, we combine ChatGPT with five deep learning models specialized in identifying discontinuous entities to enhance the model's ability to identify discontinuous entities effectively. Here is a brief introduction to these models to better understand how they address the problem of DNER:

#### 2.2.1. Transition-Based Model

The transformation-based approach proposed by Dai et al. [31] addresses the DNER problem by introducing two finer-grained entities, thereby transforming it into a nested NER problem. As illustrated in Figure 3, the original ADE entity is divided into two more specific entities: "Body Location" and "General Feeling," to solve the DNER problem.

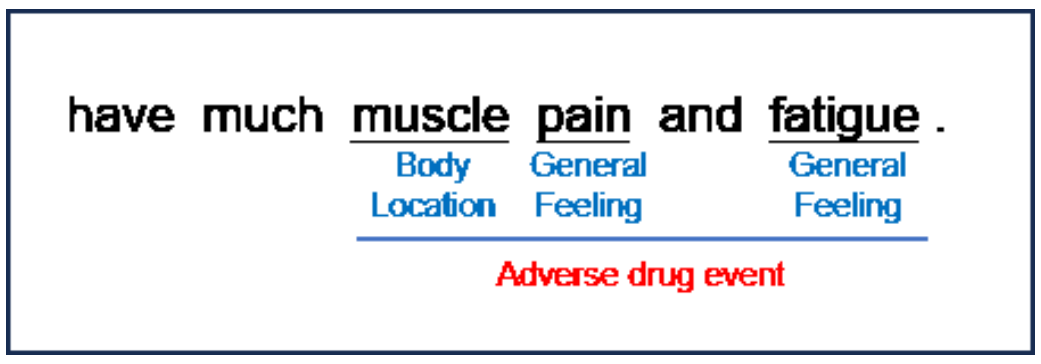

**Figure 3.** Example of the transition-based approach.

#### 2.2.2. Span-Based Model

The model proposed by Li et al. [32] is essentially a relationship extraction model. The model identifies entities by traversing all possible text spans, then performs relationship classification to determine whether the given entity pair belongs to the same named entity, two separate named entities, or entities with overlapping spans. As shown in Figure 4, this model consists of two main steps. First, it identifies entities "mitral," "leaflets," and "thickened" by traversing all possible text spans. Then, it performs relationship classification to determine that these three entities belong to the same named entity.

#### 2.2.3. Maximal Clique Discovery-Based Model

The model proposed by Wang et al. [33] is a concept in graph theory. The model is divided into two main tasks, entity extraction and edge prediction, to form the nodes and edges of the entity graph. A node graph is established for each sentence, where each node represents an entity (formed by one or more words), and edges connect nodes within the same entity. As illustrated in Figure 5, the model first identifies words that could potentially be the beginning of an ADE entity, such as "Sever joint" and "Sever" are annotated as ADE-B, and words that could be inside the entity like "pain", "should", and "upper body pain", are then annotated as ADE-I. Then, the model establishes connections between entities within the same entity through edge prediction.

#### 2.2.4. Word-Word Relation Classification Model

The $W^2NER$ model proposed by Li et al. [34] transforms the NER problem into predicting word-word relationships, and uses two custom labels to link entities together. This model effectively simulates relationships between entity words by predicting the Next-Neighboring Word (NNW) and Tail Head Word (THW) relationships. As shown in Figure 6, (a) illustrates examples of three NER scenarios: "aching in legs" (E1) is a contiguous entity, "aching in shoulders" (E2) is a discontinuous entity, and there is an overlapping entity

"aching in," while (b) demonstrates that the model converts these three NER scenarios into word-word relationship classifications. Here, Next-Neighboring-Word (NNW) indicates words that are part of the same entity (e.g., "aching" → "in"), while Tail-Head-Word (THW) signifies edges connecting the end and start of words (e.g., "legs" → "aching"). Entities are formed by NNW and THW relationships.

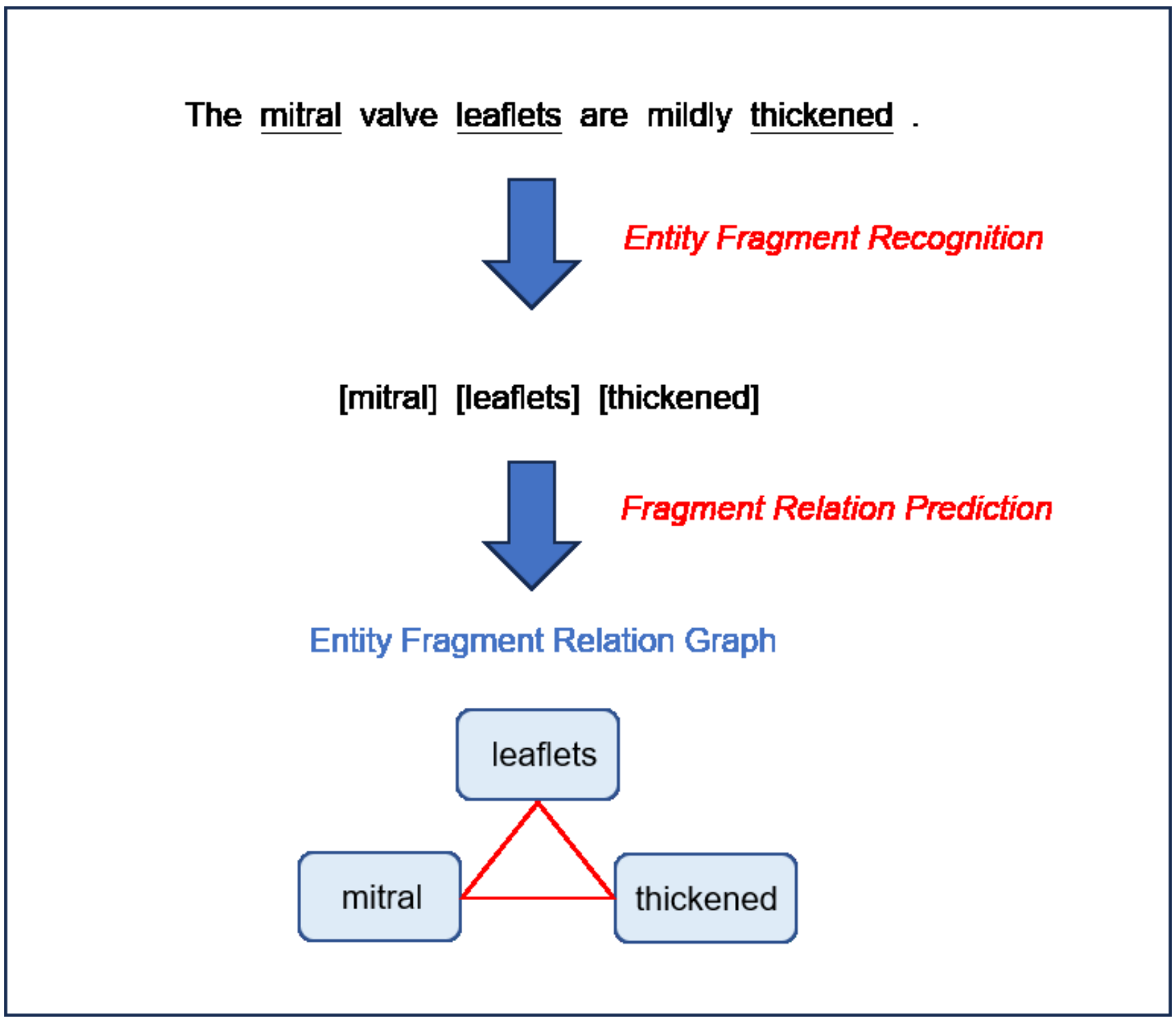

**Figure 4.** Example of a span-based model.

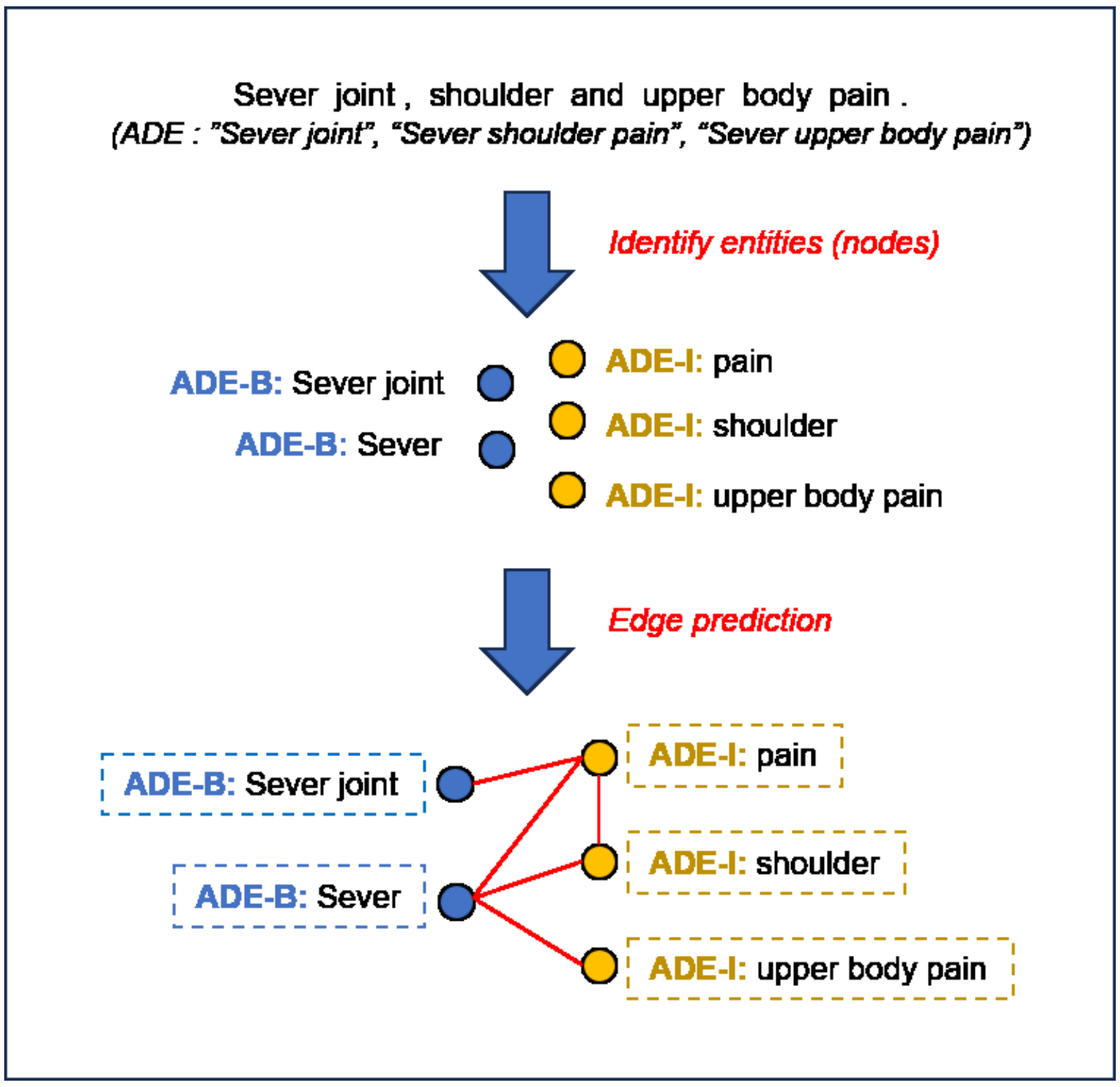

**Figure 5.** Example of the maximal clique discovery based method.

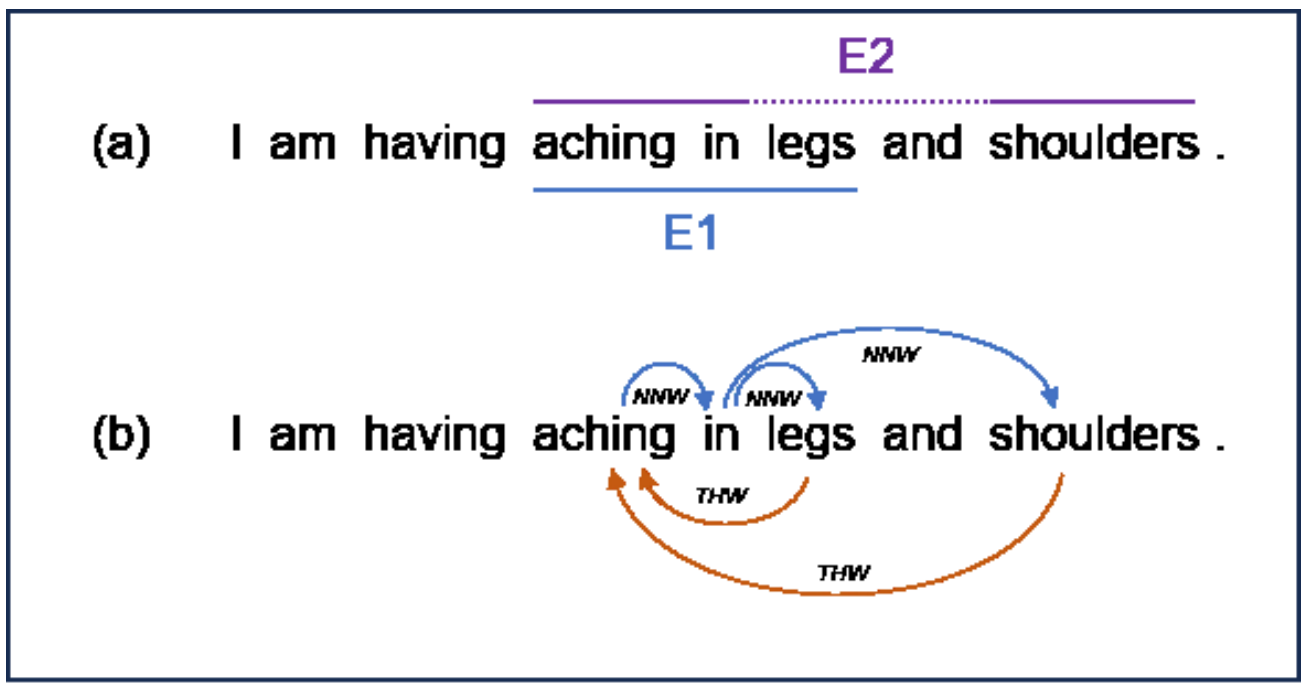

**Figure 6.** Example of the word-word relation method.

2.2.5. Tag-Oriented Enhancement Model (TOE)

The TOE (Tag-Oriented Enhancement) model proposed by Liu et al. [35] is an enhanced version of W$^2$NER model, achieving higher performance by adding two additional custom labels. As shown in Figure 7, the red relationships represent the new labels "PNW" and "HTW," which are used to enhance the W$^2$NER Model that originally only had "NNW" and "THW" labels. The model not only proposes predicting Next-Neighboring-Word (NNW) and Tail-Head-Word (THW) relationships to capture discontinuous entities, but also introduces two new relationships: Previous-Neighboring-Word (PNW) and Head-Tail-Word (HTW). This requires the model to consider not only relationships between words but also interactions between labels and words, as well as between labels themselves.

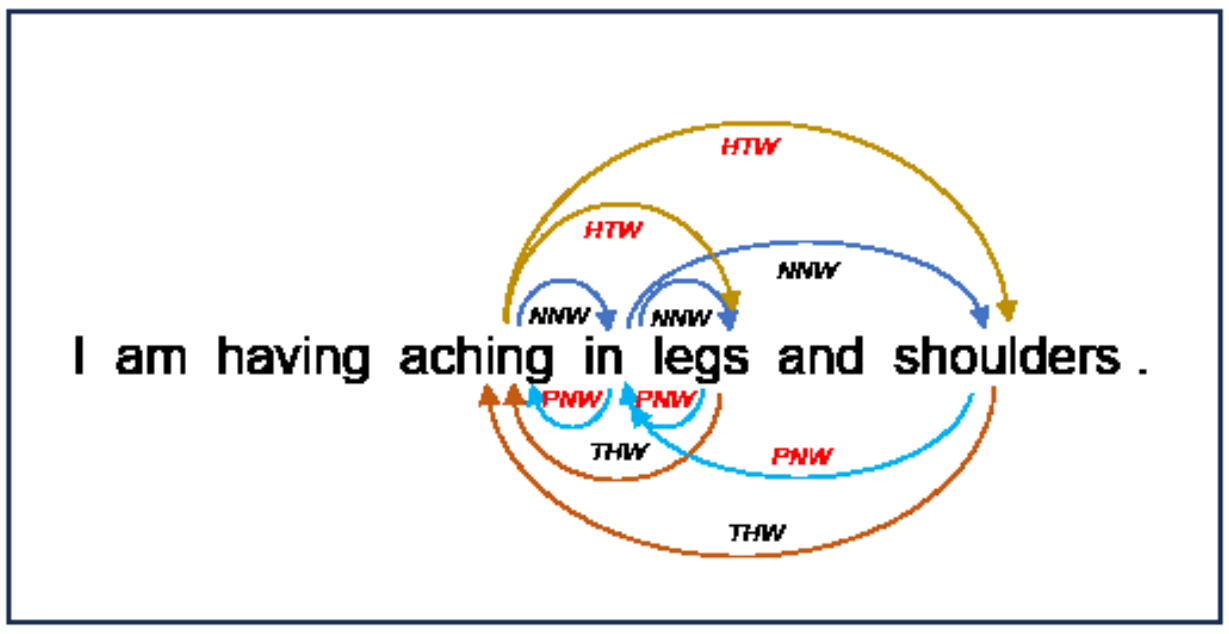

**Figure 7.** A tag-oriented enhancement paradigm of word-word relation.

## 3. Methodology

Figure 8 illustrates the overall framework of this study, aimed at leveraging the advantages of ensemble learning [36] and using ChatGPT to address the challenges of DNER in medical corpora. We employ ChatGPT as a mediator, utilizing custom prompts for ChatGPT to integrate five deep-learning models capable of handling DNER issues, thereby enhancing the model's ability to recognize discontinuous entities effectively. Additionally, for the reliability of this approach, we also explore a voting ensemble method to compare with the proposed approach of using ChatGPT as an arbitrator.

### *3.1. Data Preprocessing*

First, following the approach of [31], we preprocessed the datasets suitable for use by the DNER models. Then, we fine-tuned the data according to the input formats of each model, enabling training with the respective datasets. As shown in Figure 9, the transition-based model [31] only converts the original entity into two finer-grained entities, so only the entity position and category need to be used in the input format. The Span-based model [32] is essentially a relationship extraction model. It first predicts entities and then performs relationship classification for each entity. Therefore, "ner" needs to be used in the input format to represent entities and "relations" to represent the relationships of each entity. Maximal clique discovery-based model [33] builds a node graph for each sentence, so "entity_list" is used to represent nodes (that is, entities). "word2char_span" represents the span position within each word. The $W^2$NER [34] and TOE [35] models both identify entities by adding custom tags, so only the entity type and location are required in the input format.

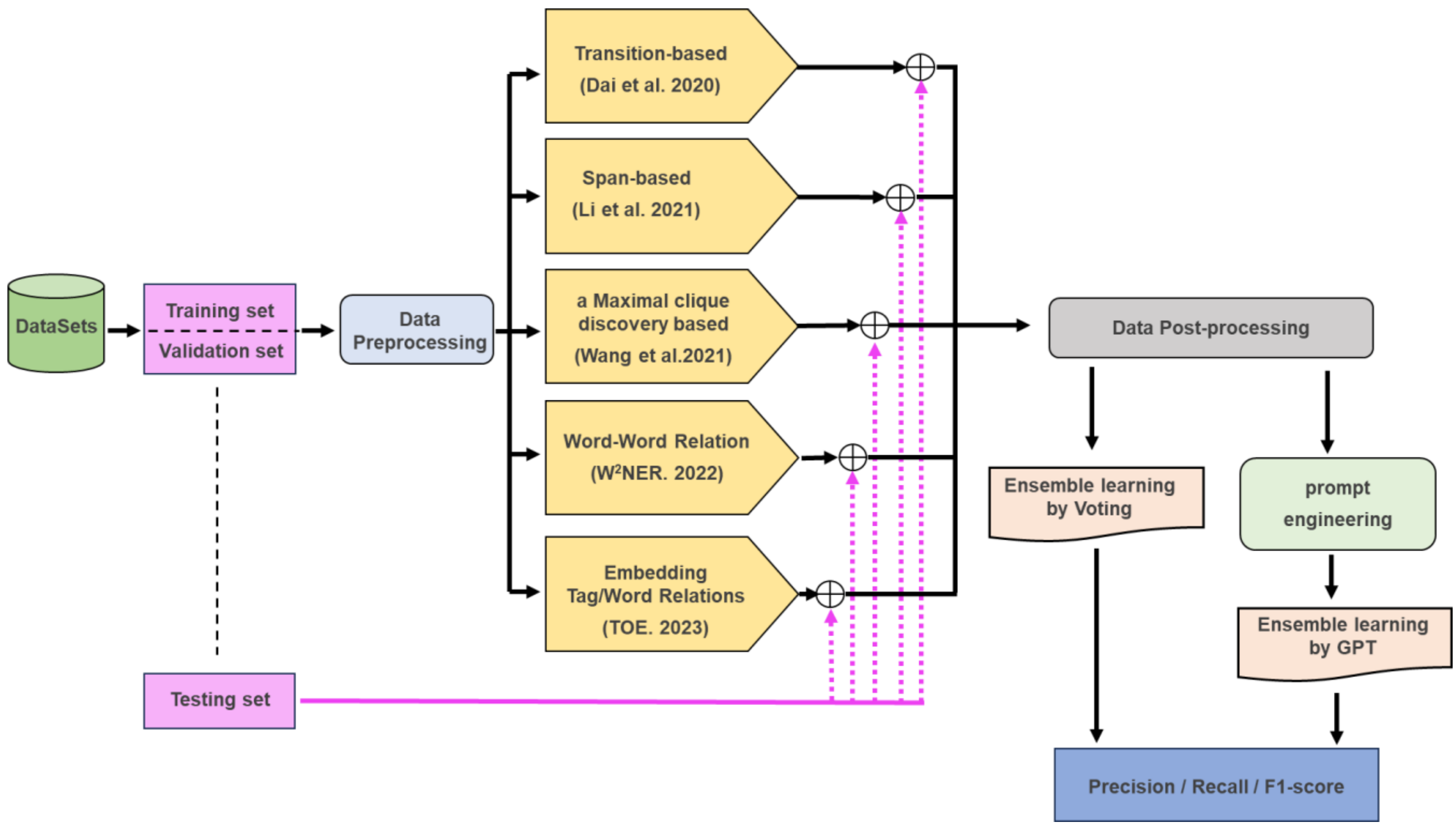

**Figure 8.** Diagram of the overall framework of this study.

*3.2. Data Post-processing*

To enable the data for learning through ensemble methods, we needed to standardize the outputs of all models into a uniform format for further evaluation of the ensemble method. Figure 10 shows an example of the proposed uniform format corresponding to the example in Figure 9. The "text" field contains the original sentence, "sentence" represents the tokenized version of the original sentence, and "entity list" includes each entity's "text" and "index" positions within the sentence. When we use majority voting in the voting ensemble method, we only need to grab the entity positions predicted by each model. However, considering ChatGPT's capabilities as a generative AI that understands text, we did not simply use entity positions like in a voting ensemble method but also included the original sentence and each entity's text, allowing ChatGPT to have a deeper understanding of the sentence content.

*3.3. Prompt Engineering*

Due to the high uncertainty in ChatGPT's responses, we applied the following prompt engineering to stabilize and constrain its answering behavior, guiding the model to generate the desired outputs. We divided the prompts into two parts. The first part is the foundational prompts, as shown in Table 1, which contains task descriptions commonly used by people when interacting with generative AI models (for example, given what role it is and what tasks it wants to handle). Annotation description is essential in the NER task, so we must also describe the entity in detail. The last is the sample description, which tells ChatGPT the input and output formats.

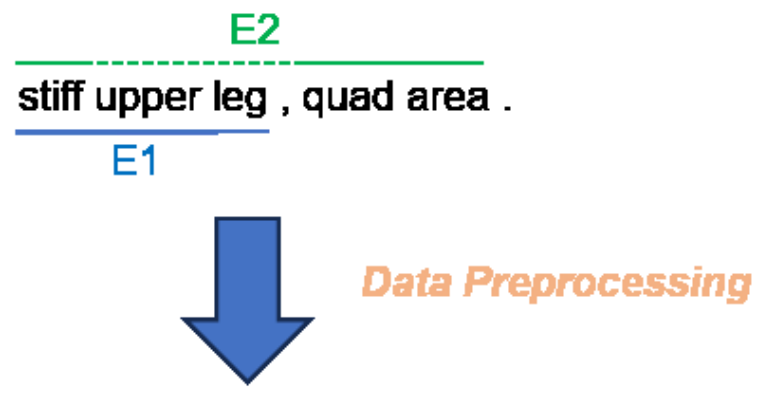

| Model name | input style |
|---|---|
| Transition-based (Dai et al. 2020) | stiff upper leg , quad area .<br>0,2 ADR\|0,0,4,5 ADR |
| Span-based (Li et al. 2021) | {"sentences": [["stiff", "upper", "leg", ",", "quad", "area", "."]],<br>"ner": [[[0, 2, "ADR"], [0, 0, "ADR"], [4, 5, "ADR"]]],<br>"relations": [[[0, 2, 0, 0, "Overlap"], [0, 0, 0, 2, "Overlap"], [0, 0, 4, 5, "Combined"], [4, 5, 0, 0, "Combined"]]],<br>"dep": [{"nodes": [[2], [2], [0, 1, 3, 5, 6], [2], [5], [4, 2], [2]]}]} |
| a Maximal clique discovery based (Wang et al.2021) | {"text": "stiff upper leg , quad area .",<br>"word_list": ["stiff", "upper", "leg", ",", "quad", "area", "."],<br>"word2char_span": [[0, 5], [6, 11], [12, 15], [16, 17], [18, 22], [23, 27], [28, 29]],<br>"entity_list": [{"text": "stiff upper leg", "type": "ADR", "char_span": [0, 15]}, {"text": "stiff quad area", "type": "ADR", "char_span": [0, 5, 18, 27]}]} |
| W²NER.2022 & TOE.2023 | {"sentence": ["stiff", "upper", "leg", ",", "quad", "area", "."],<br>"ner": [{"index": [0, 4, 5], "type": "ADR"}, {"index": [0, 1, 2], "type": "ADR"}]} |

**Figure 9.** An illustration of the data input formats of the five baseline models.

| Output style |
|---|
| {"text": "stiff upper leg , quad area .",<br>"sentence": ["stiff", "upper", "leg", ",", "quad", "area", "."],<br>"entity_list": [{"text": "stiff upper leg", "index": [0, 1, 2]},<br>{"text": "stiff quad area", "index": [0, 4, 5] }]} |

**Figure 10.** An example of the uniform format wrt the input in Figure 9 after data post-processing.

**Table 1.** An example of basic prompt engineering description.

| Basic Prompt Types | Examples |
|---|---|
| (1) Task Description | Clearly instruct it on what tasks you want it to perform and what role you want it to play.<br><br>###, e.g., You are an NER expert in the medical field who can identify side effect symptom entities and want to select the best answer from the output of five discontinuous named entity recognition models in a Health data set. |
| (2) Annotation Description | Provide comprehensive and clear descriptions of entities.<br><br>###e.g., An entity can be any adverse reaction or adverse event. These symptoms may be physical, such as nausea, vomiting, heart palpitations, headache, rash, redness, and swelling, or psychological, such as anxiety, delusions, or psychosis. |
| (3) Sample Description | Clearly explain the meaning of the input data and the desired output format.<br><br>###e.g., Input:[……..] , Output:[………] |

Unfortunately, we encountered some unsatisfied results when using ChatGPT as an arbitrator in our ensemble method. As shown in Figure 11(a), we found during execution that swapping the input order between Model 1 and Model 2 affected the subsequent results in ChatGPT. Another problem is that ChatGPT may use synonyms in the answer results. As shown in Figure 11(b), the ADR entity "redness" was output as the synonym "erythema" in ChatGPT. Therefore, we designed two specialized prompts to solve these problems. Table 2 details the design of these specified prompts.

**(a)**

| ChatGPT input | ChatGPT Output |
|---|---|
| { "entity_list_1": [{muscle pain},{muscle fatigue}] "entity_list_2": [{muscle pain}] } | {muscle pain},{muscle fatigue} |
| { "entity_list_1": [{muscle pain}] "entity_list_2": [{muscle pain},{muscle fatigue}] } | {muscle pain} |

**(b)**

| Text | ChatGPT Output |
|---|---|
| "My arms are redness." | {arms erythema } |
| {"text": "My arms are redness.", "sentence": ["My", "arms", "are", "redness", "."] } | {arms redness } |

**Figure 11.** An example of the problems encountered when using basic prompts for ChatGPT's output. **(a)** Input ordering problem; **(b)** Output synonym problem.

**Table 2.** Description of specialized prompt engineering.

| Special Prompt Types | Guideline |
|---|---|
| (1) Input ordering problem | We directed GhatGPT to act as an arbitrator between "entity lists" to ensure judgments are not influenced by the output order of each model in the "entity list." |
| (2) Output synonym problem | We added "sentence" in the tokenization process of the model, instructing ChatGPT's output to select entities based on permutations from the "sentence," thereby allowing for multiple choices. |

## 4. Evaluation

All experiments were performed on a PC equipped with the following specifications: an Intel Core i5-12400 CPU, 32GB RAM, a 1TB SSD hard disk, and an NVIDIA GeForce RTX 3070-Ti with 8GB VRAM graphics card, running on Windows 10, and all software was implemented in Python.

*4.1. Data Sets*

To underscore the reliability of our proposed method, we applied preprocessing to three benchmark biomedical datasets of CADEC, ShARe13, and ShARe14, as previously done by Dai et al. [31].

CADEC [23] is a richly annotated corpus containing medical forum posts where patients report adverse drug events. The texts in this corpus are mostly written in informal language and frequently diverge from standard English grammar and punctuation norms. Annotation quality is maintained through the use of guidelines, a multi-phase annotation process, inter-annotator agreement measurements, and final reviews by clinical terminologists. This corpus is valuable for research in extracting information from social media or text mining to detect potential adverse drug reactions directly from patient reports.

ShaRe13 [24] and ShaRe14 [25] are datasets belonging to a shared task. In the ShaRe13, the laboratory includes three tasks: Task 1 involves disease identification and standardization (1a and 1b), and Task 2 involves standardizing medical term abbreviations and acronyms. Task 3 focuses on information retrieval. In this study, we used Task 1 to evaluate NER performance. The ShaRe14 also comprises three tasks: Task 1 focuses on interactive visualization and exploration of electronic health records, Task 2 involves information extraction from clinical texts, and Task 3 is dedicated to user-centered health information retrieval. Our experiments used the Task 2 dataset of ShaRe14 to assess NER performance.

Table 3 presents the descriptive statistics of the datasets, including document counts, sentence counts, token counts, and entity counts. Here, Disc.E represents the number of discontinuous named entities comprising approximately 10% of the total entities.

**Table 3.** The descriptive statistics of the datasets.

| | **CADEC** | **ShARe13** | **ShARe14** |
|---|---|---|---|
| **Text type** | Online posts | Clinical notes | Clinical notes |
| **Entity type** | ADE | Disorder | Disorder |
| **Documents** | 1,250 | 299 | 431 |
| **Sentences** | 7,597 | 18,767 | 34,618 |
| **Tokens** | 122,938 | 278,942 | 522,355 |
| **Entities** | 6,318 | 11,148 | 19,073 |
| **Disc.E** | 679 | 1,088 | 1,656 |

*4.2. Performance Metrics*

In NER tasks, Precision, Recall, and F1-Score are the standard metrics for evaluation. These metrics effectively assess the performance of NER systems. Calculating accuracy is more difficult because it is challenging to determine the exact True Negative (TN) value. This is because the main goal of NER is to identify entities in the text rather than performing binary classification (entity/non-entity) for each word. Therefore, we selected three metrics to evaluate our experiments: precision, recall, and F1-score, all defined by confusion matrices.

*4.3. Results and Discussion*

We compared our method with five deep learning-based methods, as well as GPT-3.5 and GPT-4. The test results are shown in Table 4, where highlighted in bold indicate the best results for each metric, and underlined values indicate the second-best results. The first five rows represent the five baseline models, the sixth and seventh rows represent the generative AI models GPT-3.5 and GPT-4, and the last two rows represent ensemble learning methods: a simple majority voting ensemble method and an ensemble method using ChatGPT as an arbitrator.

The results show that our ChatGPT-coordinated ensemble algorithm outperforms five baseline models, generative AI models, and voting ensemble methods in terms of F1 score. Moreover, across the three benchmark datasets, out method elevates the state-of-the-art (SOTA), i.e., TOE [35] in the baseline models, results by approximately 1.13%, 0.54%, and 0.67%. Compared to voting ensemble methods, our approach showed improvements of about 0.63%, 0.32%, and 0.09%. Additionally, when individually compared to GPT-3.5 and GPT-4, ChatGPT-coordinated ensemble achieves average improvements of about 7.42%, 0.89%, and 0.54%.

In Figure 12, the critical difference diagram visually highlights that ensemble learning methods consistently outperformed baseline models and generative AI models in terms of precision and F1-score across the three benchmark datasets. We observed that generative AI models (GPT-4) excelled in the recall evaluation metric but exhibited lower precision than other models.

**Table 4.** Experimental results obtained in datasets of CADEC, ShARe13, and ShARe14.

| Model | | CADEC | | | ShARe13 | | | ShARe14 | | |
|---|---|---|---|---|---|---|---|---|---|---|
| | | P | R | F1 | P | R | F1 | P | R | F1 |
| **transition-based** | Dai et al. [31] | 67.30 | 67.50 | 67.40 | 80.44 | 74.81 | 77.52 | 76.18 | 81.20 | 78.60 |
| **Span-based** | Li et al. [32] | 68.50 | 69.90 | 69.19 | 83.44 | 79.81 | 81.62 | **81.75** | 81.57 | 81.17 |
| **Grid Tagging** | Wang et al.[33] | 69.60 | 70.85 | 70.22 | 83.20 | 78.60 | 80.83 | 78.69 | 82.15 | 80.39 |
| | W²NER [34] | 72.59 | 70.15 | 71.35 | 83.97 | 79.08 | 81.45 | 79.82 | 82.11 | 80.95 |
| | TOE [35] | 75.36 | 69.16 | 72.13 | 83.78 | 79.52 | 81.59 | 80.78 | 81.57 | 81.21 |
| **Gen AI** | GPT-3.5 | 60.55 | 71.85 | 65.72 | 79.80 | 82.17 | 80.97 | 79.82 | **82.41** | 81.09 |
| | GPT-4 | 62.43 | **79.90** | 70.09 | 80.14 | **83.22** | 81.64 | 80.80 | 82.35 | 81.55 |
| **Ensemble learning** | use voting | **76.55** | 68.83 | 72.49 | **84.62** | 79.11 | 81.77 | 81.46 | 81.99 | 81.69 |
| | use GPT-4 | 76.03 | 70.11 | **72.95** | 82.52 | 81.55 | **82.03** | 81.51 | 82.12 | **81.76** |

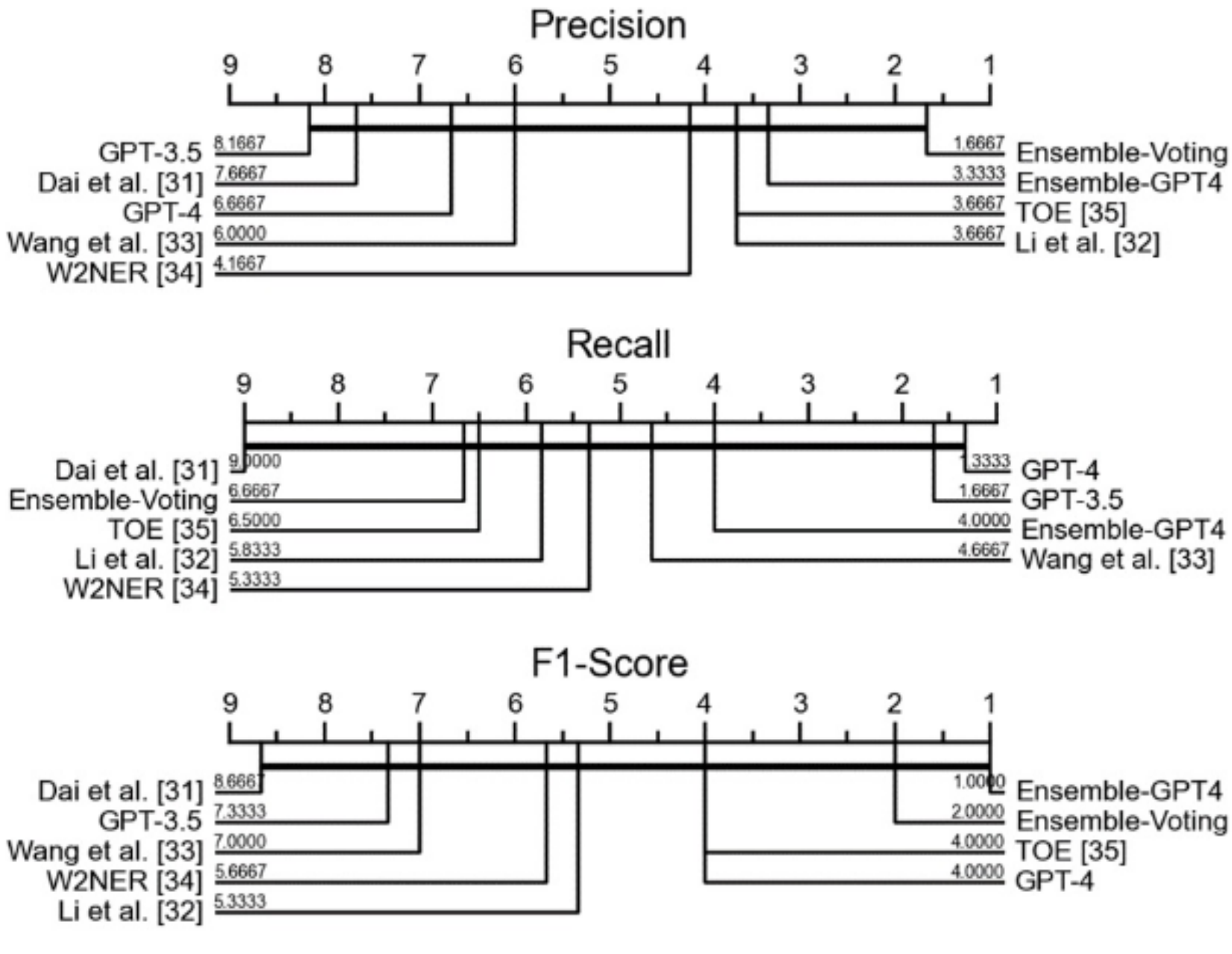

**Figure 12.** Performance comparison vis critical difference diagram.

In summary, we have the following observations:

1. The ensemble method we proposed, using ChatGPT as an arbitrator, effectively enhances the performance of individual models on the DNER problem.
2. Conventional voting ensemble also effectively improves the performance of DNER models, but their effectiveness is slightly lower than our proposed method of using ChatGPT as an arbitrator.
3. In addressing the DNER problem, ChatGPT shows a notably high recall but low precision, which could be attributed to the hallucination issue typical in generative AI models. This is because we did not restrict ChatGPT's response scope, leading to a high volume of responses. Consequently, this significantly reduces False Negatives while substantially increasing False Positives.

## 5. Conclusions

Many research institutions have recently employed various deep-learning models to address the DNER problem. Nevertheless, we have not found relevant studies using ensemble learning to tackle DNER issues. We explored two ensemble learning methods to investigate whether they can enhance the performance of individual DNER models. One method is the majority voting ensemble approach, while the other is our novel method, using ChatGPT as an arbitrator to combine outputs from other deep learning models.

We have conducted comprehensive experiments on three benchmark medical datasets. The results demonstrate that our proposed approach of ChatGPT-coordinated ensemble algorithm outperforms other individual deep learning models, ChatGPT itself, or the voting ensemble algorithm. In summary, our study demonstrated the potential of using ChatGPT as a coordinator to shape a better ensemble-based DNER model.

**Funding:** This research was partially supported by National Science and Technology Council of Taiwan, grant number 111-2221-E-390 -011 -MY2.

**Data Availability Statement:** The programs and data presented in the study are openly available in the GitHub repository at https://github.com/NUKCILAB/On_Fusing_ChatGPT_and_Ensemble_Learning_in_Discontinuous_Named_Entity_Recognition_in_Health_Corpora (accessed on 1 July 2025).

**Conflicts of Interest:** The authors declare no conflicts of interest.